# Human-centric Computing and Information Sciences

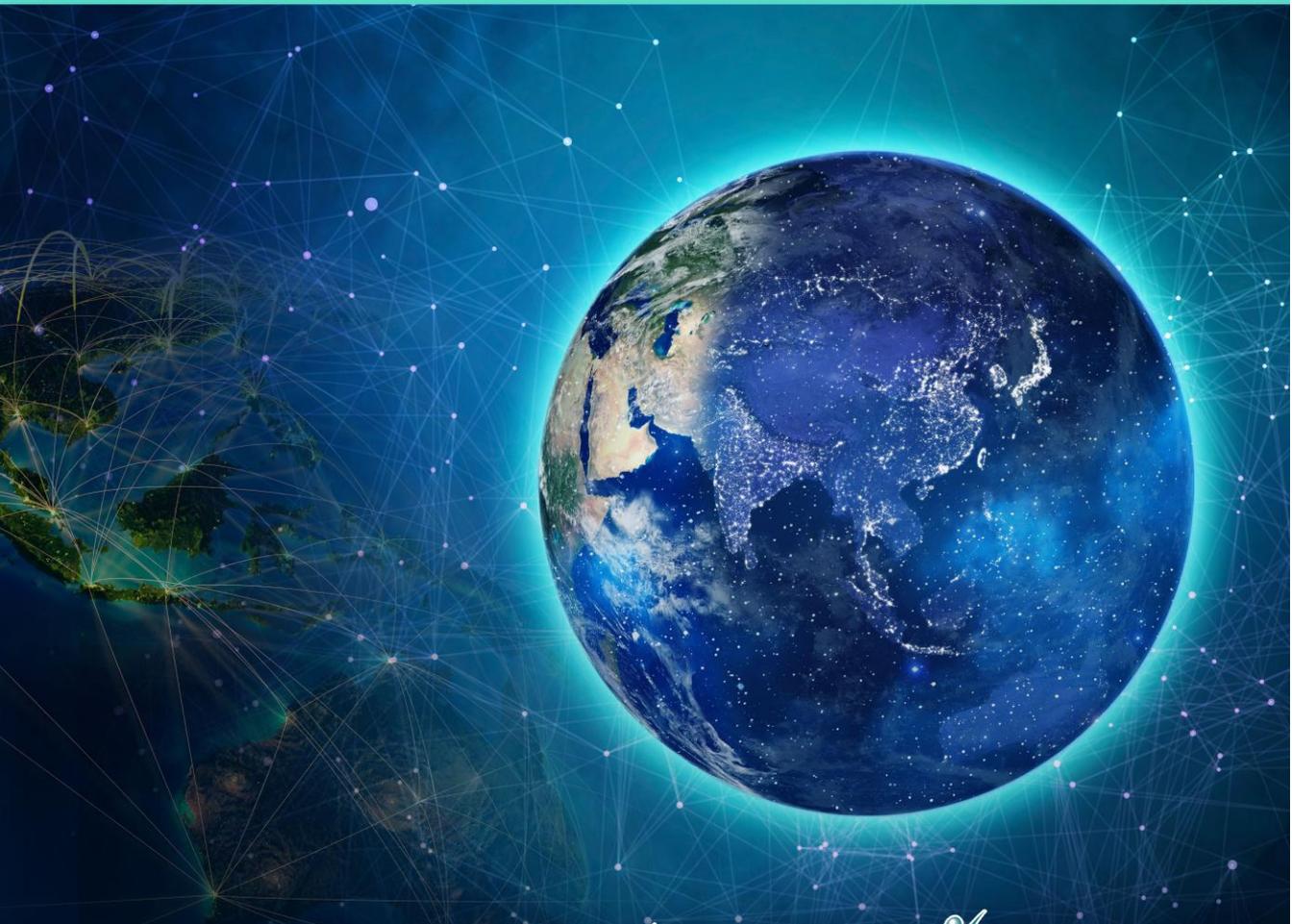

# Deliberative Context-Aware Ambient Intelligence System for Assisted Living Homes

Mohannad Babli*, Jaime A Rincon, Eva Onaindia, Carlos Carrascosa, and Vicente Julian

## Abstract
Monitoring wellbeing and stress is one of the problems covered by ambient intelligence, as stress is a significant cause of human illnesses directly affecting our emotional state. The primary aim was to propose a deliberation architecture for an ambient intelligence healthcare application. The architecture provides a plan for comforting stressed seniors suffering from negative emotions in an assisted living home and executes the plan considering the environment's dynamic nature. Literature was reviewed to identify the convergence between deliberation and ambient intelligence and the latter's latest healthcare trends. A deliberation function was designed to achieve context-aware dynamic human-robot interaction, perception, planning capabilities, reactivity, and context-awareness with regard to the environment. A number of experimental case studies in a simulated assisted living home scenario were conducted to demonstrate the approach's behavior and validity. The proposed methods were validated to show classification accuracy. The validation showed that the deliberation function has effectively achieved its deliberative objectives.

## 1. Introduction

Population aging has become a global phenomenon, bringing challenges and increasing the demand for healthcare solutions. Ambient intelligence (AmI) refers to the intelligent software that supports people in their daily lives by sensibly assisting them [1]. AmI has provenuseful for healthcare and senior citizens, and it has been used in a broad spectrum of related applications, such as monitoring seniors and alerting caregivers, recognizing seniors'activities, detecting Alzheimerearly, managing stress, and empowering people living with the early stages of dementia with autonomy [2–4].

The ability to reason and the inclusion of cognitive tasks are increasingly demanded in AmI to deal with stress-related disorders and other health issues. At least one of the following features is highlighted for the successful application of AmI in healthcare: (1) context-awareness; (2) learning; (3) reactivity and proactivity; (4) reasoning with dynamic environments; and (5) reasoning with time. These five features, along with human interaction, converge towardintelligent deliberation notion.

Context-awareness is the capability of a system to understand the current situation (environment context), track changes, and relate this knowledge to the system to produce proactive reactions [1]. Learning how the senior feels (personal context), planning what the senior needs [5], and reasoning about time to maintain an orderly schedule of the senior activities are also crucial functionalities.

Whether in open or closed well-structured environmentssuch as assisted living (AL), autonomously performing tasks and interacting with humans require deliberation capabilities to adapt to circumstances. In this work, we adopt the deliberation definition of Ingrand and Ghallab [6], which conceptually distinguishes six functions required for successful deliberation: planning, acting, perceiving, monitoring, learning, and goal reasoning. The convergence between the desired features in AmI and deliberation is natural due to the technologies being used. The combination of learning and perception provides context-awareness, and the combination of planning, acting, and monitoring provides reactivity/proactivity to deal with incomplete knowledge and dynamic environments.

Despite being promising, current AmI solutions for seniors' healthcare lack a balanced, integral deliberative function. Planning-focused approaches are useful for proactivity and reactivity but fall short on semantic knowledge reasoning and vice versa. Methods that focus on cognition, context-awareness, learning, activities recognition, and sensors tend to lack planning capabilities[7], where the automation module is a communication organizer.

This paper's scope is on human-centric technologies that require close alignment of the seniors and the artificial intelligence (AI) interacting with the environment to provide context-awareness. To tackle the challenges above, this contribution proposes a deliberation architecture for an AmI healthcare application that provides comfort to stressed seniors suffering from negative emotions in an AL environment. The deliberation architecture makes a balanced synthesis of data and knowledge, learning, perception, dynamic goal generation, planning, acting, and monitoring functionalities required to detect, address, and follow up on the senior status evolution. The rest of the paper is organized as follows:in the related work section, we reviewthe importance of the various components to achieve deliberation in AmI and the most advanced methods for detecting stress and recognizing emotions; in the concept design section, we describe our AL environment and its components and subsequentlyexplain the planning task representing the suggested concept design; we provide the system schema and briefly explain our deliberative function architecture and its five layers in the proposed methodology section; afterward, we describe the details of the deliberation function; we demonstrate the behavior and validity of the approach and the proposed methods in the validation section; then, we outline areas for improvement and other implications in the discussion section;finally, we conclude the paper in the last section.

## 2. Related Work

Context-awareness enhances AmI applications for elderly healthcare and stress management in particular [8]. They enable such systems to recognize seniors, their status, and the surrounding environment, allowing richer reasoning and better-informed decisions and actions. Perception, also known as observing, is a critical component in deliberation [6]. IoT advances, and the scarce availability of human assistance has led to AL. Such environments may be equipped with IoT sensors for determining the availability and location of the objects and for sensing anomalies such as when a senior suffers stress or falls, making them appealing for seniors' healthcare applications.

Planning, acting, monitoring, and reasoning are crucial for deliberation in AmI healthcare systems to handle time as well as the dynamic environment when a plan is executed. The AmI homecare system for diabetic patients [9] utilizes a distributed hierarchal task network to coordinate a plan, monitors its execution and the environment to adjust the room temperature, suggests insulin injection, or alerts the user on the need for medical attention. COACH [10] relies on planning and computer vision to support aging-in-place safety. The framework [5] for domestic healthcare robots generates a goal dynamically according to a cognition layer and uses a task planner to plan actions. The generalized argumentation framework [11] takes adaptive decisions and reasons with the evolution of inhabitants' preferences overtime on lighting, healthy eating, and leisure inside a smart home. The framework in [12] utilizes temporal reasoning with a rule-based system to recognize hazardous situations in a smart home environment, facilitating decision making that resolves the anomaly and returns the environment to a safe state.

Ontologies are helpful informally definingthe terms and relationships representing the vocabulary of a domain and its context, allowing the reuse of a richer, fine-grained knowledge to assist in the performance of everyday tasks and promoting context-awarenesstoward everyday environments [13].

Machine learning has been popular in recent years for context recognition, recognizing human activities and emotionsand observing environment objects [14].

A variation in stress levels directly affects the emotional state and negatively influences physical and mental health. A recent study [2] shows that the top problem-related category for AL is negative emotions. Stress is detected through a variety of bio-signals such as electroencephalography (EEG), photoplethysmography (PPG)and electrocardiogram (ECG), and facial expressions analysis, which is contactless, allowingdetecting a larger spectrum of emotions [15].

## 3. Concept Design of the AL Environment

Helping clients understand self-care is vital in relapse prevention and recovery facilities, AL homes, or nursing homes. Emotions such as terror, sadness, and boredom can also occur in response to trauma. If taken into consideration, simple self-care indicators could prevent accidents or improve fallprevention plans by paying attention to hydration. To this extent, we aim to design a plan for providing hydration, cheering, or entertainmentwhen a stressed senior feels terrified, sad, or bored, respectively. Theplan is then executed, taking into consideration the dynamic nature of the AmI environment.

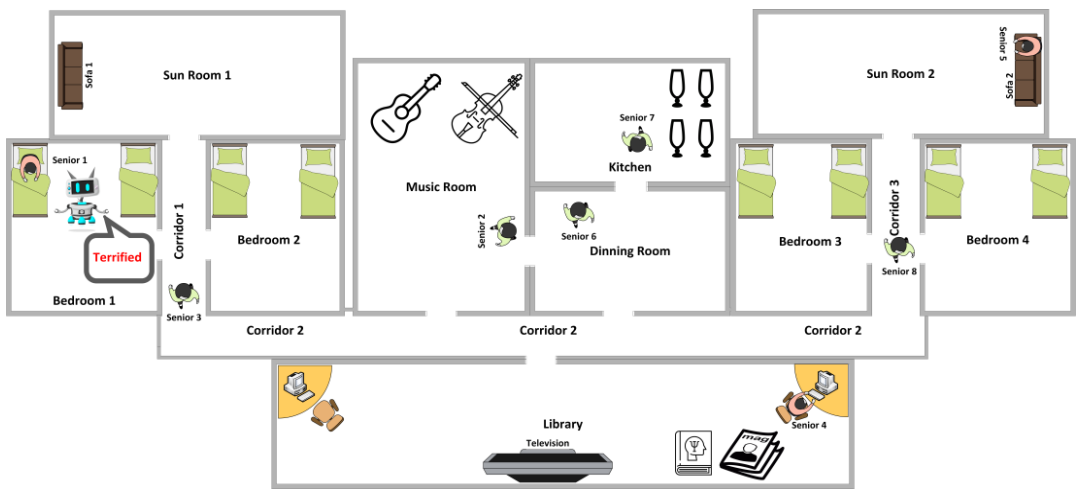

**Fig. 1.** Concept design of an AL home using the system.

Fig. 1 shows the AL environment's concept design, whereas the planning task's description is explained in the next section. The environment consists of four bedrooms, two sunrooms, a music room, a dining room, a kitchen, a library, three corridors that link the spaces mentioned above, a robot, and several senior citizens wearing IoT wristbands. The environment features an additional set of IoT objects of different categories (types) distributed in the AL home.Drinking glasses fall under the category of drinking vessels in the kitchen. A violin and a guitar fall under the category of music instruments in the music room.Entertainment objects in the library are to be given, such as magazines and books, or to be suggested for use, such as PCs and TVs. There are two types of actions by the robot in the environment:doing an activity (moving between rooms, carrying an object, giving an object, filling a drinking vessel) and suggesting an activity. Due to the environment's dynamic nature, existing IoT objects could become unavailable (being unexpectedly used by other seniors or due to a malfunction). Additionally, new non-IoT objects that have no connection to the grid and which were not previously introduced in the environment may also be encountered (added by the caretakers) during the plan's execution and could be used to repair execution failures.

## 4. Planning Task of the AL Concept Design

We represent the AL environment's concept design as a temporal planning task using version 2.2 of the popular planning domain definition language (PDDL). Temporal planning involves selecting and organizing actions to satisfy the goals and assigning to each of these actions its start time and duration. A temporal planning task is described by the planning domain and the planning problem. The planning problem variesdepending on the particular problem to be solved. In contrast, the planning domain is a fixed description of what the planner can do to achieve a set of goal conditions starting from an initial state. The planning domain characterizes the planning task behavior, comprises the model (shown in Fig. 2) of the AL environment, and consists of the following:

```
(:types
    robot senior location drinking_vessel making_music - object
    entertainment - object
    solving reading communication_device - entertainment
    drinking_glass glass_bottle - drinking_vessel
    violin guitar - making_music
    puzzle yoyo - solving
    magazine book  - reading
    tv desktop - communication_device
    dinning_room kitchen music_room library bedroom - location
    corridor sunroom - location)

(:predicates
;; locatable is at a location
(be ?locatable (either robot senior drinking_vessel
    making_music entertainment) ?loc - location)
;; a link exists between two location
(path ?loc1 - location ?loc2 - location)
;; time-window for soothing terrified
(hydration_time ?senior - senior)
;; the senior has fear (terrified)
(terrified ?senior - senior)
;; the senior is sad
(sad ?senior - senior)
;; the senior is bored
(bored ?senior - senior)
;; fear is soothed
(soothed ?senior - senior)
;; sadness is soothed and the person is cheerful
(cheerful ?senior - senior)
;; boredom is soothed and the person is entertained
(entertained ?senior - senior)
;; availability of an object
(available ?object - (either drinking_vessel
    making_music entertainment))
;; status of a drinking vessel
(filled ?item - drinking_vessel)
;; the robot is carrying an object
(holding ?robot - robot ?item - (either drinking_vessel
    making_music solving reading))
;; an object has already been given to a senior
(given ?item - (either drinking_vessel making_music
    solving reading) ?senior - senior)
;; a suggestion has already been given to a senior
(suggested ?item - (either tv desktop) ?senior - senior)

(:durative-actions heads
;; movement of robot between two connected locations
(move (?robot - robot ?loc1 - location ?loc2 - location))
;; the robot carries an item from a location
(carry (?robot - robot ?item - (either drinking_vessel
    making_music solving reading) ?location - location))
;; the robot fills up a drinking vessel in the kitchen
(fill (?robot - robot ?item - drinking_vessel
                    ?location - kitchen))
;; the robot gives a senior an item
(give (?robot - robot ?senior - senior ?item - (either
    drinking_vessel making_music solving reading)
    ?location - location))
;; senior received soothing for fear state
(soothed_received (?senior - senior
    ?item - drinking_vessel))
;; the senior received the cheering up to soothe sadness
(cheer_received (?senior - senior ?item - making_music))
;; the senior received the entertainment to soothe boredom
(entertainment1_received (?senior - senior ?item - (either
    solving reading)))
;; the robot suggest a different course of entertainment
(suggest_entertainment2 (?robot - robot ?senior - senior
    ?item - (either tv desktop) ?location - location))
;; the senior has received the entertainment suggestion
(entertainment2_suggested senior (?senior - senior
    ?item - (either tv desktop)))
```

**Fig. 2.** Set of types, Boolean variables, and head parts of the set of actions schemas.

1) set of object types such as `robot`, `senior`, `location`, `drinking_vessel`, `making_music` and `entertainment`.
2) set of Boolean variables such as `(path ?loc1 ?loc2 - location)` to indicate that two locations are linked.
3) set of function variables mapping to a numerical rational value, which we strictly use for actions' durations or costs, such as `(moving_time ?loc1 ?loc2 - location)`, representing the moving time between two locations.
4) action schemas representing operations that can be performed, e.g., `(move ?robot - robot ?loc1 ?loc2 - location)`..., which allows the robot to move between locations.

The planning domain is designed to penalize the actions of suggestion (action costs) in the metric to favor actions that include providing an object to seniors instead of merely suggesting an activity. Suggesting an activity is left as a last resort when there are no available objects to comfort the senior.

On the other hand, the planning problem represents a particular problem instance to be solved, specified by the initial state and the goal. It follows the basic planning problem's information: three corridors, `corridor1...3`; four bedrooms, `bedroom1...4`; one kitchen, `kitchen1`; one music room, `music_room1`; one library, `library1`; one dining room, `dining_room1`; two sunrooms, `sunroom1` and `sunroom2`. The paths between locations (layout), e.g., `(path sunroom1 corridor1)`. Eight seniors, `senior1...8`; one robot, `robot1`; eight glasses, `glass_id01...08`; one violin `violin_id01` and one guitar `guitar_id01`;two desktops `pc_id01` and `pc_id02`and one television `tv_id01`;one book `book_id01` and one magazine `magazine_id01`. The planning problem's information specifies the initial state—the locations of the environment objects, the seniors and the robot, and the availability of objects sensed from IoT objects. The goal condition will be dynamically generatedas explained in the following section.

## 5. Proposed Methodology

To achieve our aim, we designed the deliberation function schema (sketched in Fig.3). The function consists of five layers. The offline layer (layer 0 learning and knowledge augmentation) consists of four modules: learning stress, learning emotions, types augmentation, and learning types. On the other hand, the online layers arethe perception layer (layer 1), the goal generation layer (layer 2), the planning layer (layer 3), and the acting and monitoring layer (layer 4).

The input to our systemis shown in Fig. 3 in Roman numerals. The input consists of the following: (I) a multimodal database for stress classification; (II) a dataset with 4,900 images of human facial expressions; (III) the basic planning domain model follows our AL environment concept design and a repository of several semi-cooperative agents' remote partial planning tasks;(IV) the data related to the IoT objects (e.g., glasses, violins, guitars, magazines, books, televisions, and computers) and IoT wristbands. Details about the input are found in the system details section.

For context-aware human-robot interaction towardsseniors, the system can determine online whether the senior wearing an IoT wristbandis suffering from stress. Then the system recognizes the stressed senior negative emotion and dynamically generates a goal that corresponds to the negative emotion.This is achieved by the perception layer and the goal generation layer. The system utilizes the learning stress and emotions modules to train the system offline and prepare to perform them online.

The system utilizesthe three layers below in perception, planning, acting, and monitoringto achieve perceptiveness, planning, and reactivity when handling failures and context-awareness towardthe environment online. The system uses types augmentation and learning types modules to train it offline to be prepared to perform these tasks online. The deliberation layers are:

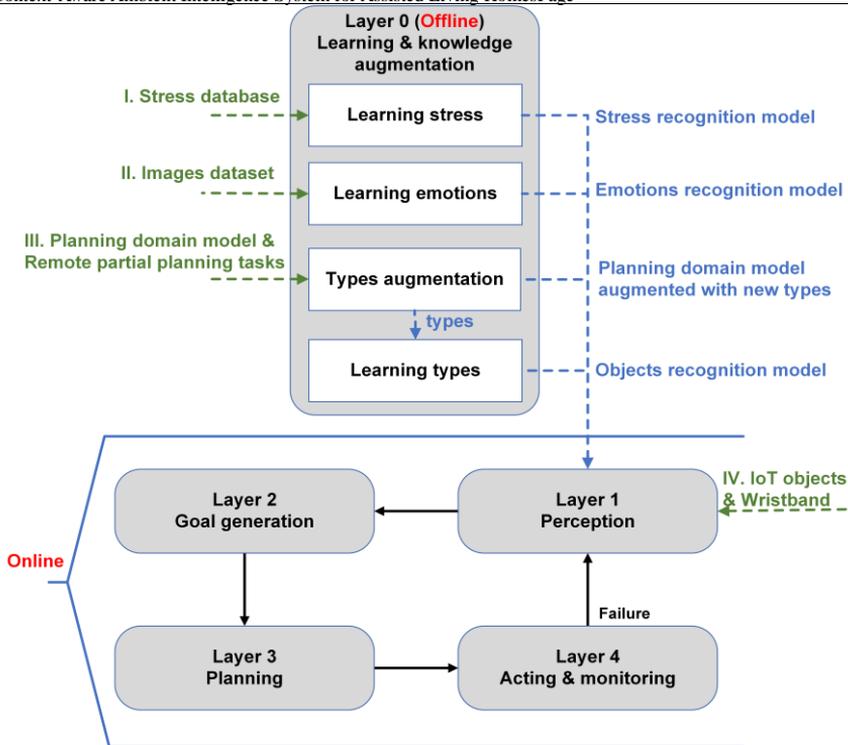

**Fig. 3.** Schematic view of the deliberation function system architecture.

1) Layer 0: learning and knowledge augmentation, executed offline to produce three classification models for recognizingstress, emotions, and objects. Layer 0 consists of four modules:
   a) Learning stress: processes the multimodal database for stress and trains the system to prepare it to detect stress by providing the stress classifier model.
   b) Learning emotions: processes the dataset of images of facial expressions and trains the system to prepare it to recognize emotions by providing the emotions classifier model.
   c) Types augmentation: is a cognitive semantic module that dynamically augments the predefined list of object types of the planning task with relevant new object types; to be context-aware towardthe environment and the task being performed and reason with incomplete knowledge, thereby boosting the system's autonomy and context-awareness.
   d) Learning types: trains the system on the augmented set of objects types to prepare it to recognize objects of the relevant types by providing the objects classifier model.
2) Layer 1: perception provides our system with the critical capability of observing by:
   a) Reading the seniors' biosignals using the IoT wristbands and sending these signals to the service in charge of the analysis using the stress classifier for stress detection; then, recognizing the negative emotion of the stressed senior using the emotions classifier model.
   b) Sensing the IoT objects' values characterizingthe environment's actual state, e.g., the seniors' locations, the robot's location, and the objects' availability.
   c) If a failure occurs, perception provides reactivity. During the plan's execution,the robot may encounter new non-IoT objects of the relevant types not predefined in the model. The robot uses the objects classifier to recognize such objects dynamically, integrates them into the planning problem, and uses them to repair failures.
3) Layer 2: goal generation is responsible for dynamically generating the goal associated with resolving the recognized negative emotion.
4) Layer 3: planning is responsible for:

   a) formulating the planning problem.
   b) reformulating the planning problem (in case of failure) starting from the state in which the failure occurred, considering the new observed state and the newly integrated objects.
   c) calling an external planner to synthesize a plan of actions to be executed by the robot to achieve the goal and comfort the senior suffering from the negative emotion.
5) Layer 4: acting and monitoring are responsible for executing the plan, monitoring action,and detecting failures that may occur when an IoT object used in the plan becomes unavailable.

The logic behind recognizing the emotion in addition to sensing stress for goal generation is that the IoT wristband is prone to noise; thus, affectingthe stress classification processas shown in the validation section. The technical details of the deliberation function are shown in the next section.

## 6. System Details

This section details the proposed deliberation function. It describes the wristband hardware used in the perception layer, stress learning and detection, emotions learning and recognition, semantic knowledge augmentation, objects types learning, planning, acting, and monitoring.

## 6.1 StressDetection and Emotions Recognition

The aim is to train the system to prepare it to detect stress andrecognizenegative emotions to generate the goal corresponding to comforting the negative emotiondynamically.

The wristband we used has a PPG sensor,and it was designed using an ESP32 Oled Lora TTGO LoRa32 development board. The TTGO is responsible for acquiring the PPG sensor signal. The signal is processed and sent through the Wi-Fi module to the web servicein charge of signals analysis and detection of stress levels. The web service is programmed using the Flask web application framework.

Since our system only incorporates a single PPG sensor, and it is possible to detect stress levels using the heart rate (HR) and the interval between the PPG signal peaks (less intrusive than ECG), a new database was created from the database [3]. This new database has only two inputs and one output (stressed or unstressed). The new database was divided into two parts to carry out the training process: 80% for training and 20% for testing. A k-fold cross-validation statistical method [16] was performed to determine independently the partitions between training and test data and estimate the learning model's ability. The best results were obtained with a k=4.

Our system incorporates a convolutional neural network (CNN)of three convolutional layers [3, 32, 32]to classify human emotions. To perform the training and the test, we used the Karolinska Directed Emotional Faces (KDEF)database [17].KDEF has 4,900 images of facial expressions for 70 actors (35 men and 35 women) showing 7 different emotions: terror, anger, disgust, happiness, neutral, sadness, and surprise. Since the KDEF database does not include the boredom emotion, a new class has been created using the UMB-DBdatabase [18]. The new dataset was divided into two parts: 80% of the images were used for training, and 20% were used to test. The images were re-sized to 48×48 pixels. The CNN hyperparameters are as follows: 0.01 for the l2 regularization or l2-penalty, [32, 64, 128, 64, 32]as hidden layers, 0.2 for the dropout rate, val_loss for the monitor, and 10 for the min delta. An example of emotion detection is shown in Fig.4 for detecting the emotion terrified. Once the emotion of a stressed senior is recognized, the system dynamically generates the goal corresponding to the emotion `(soothed senior)` if terrified, `(cheerful senior)` if sad, and `(entertained senior)` if bored. The variable `(soothed senior)` represents the goal that needs to be accomplished by the plan computed by the planning solver to set the variable `(soothed senior1)` to true.

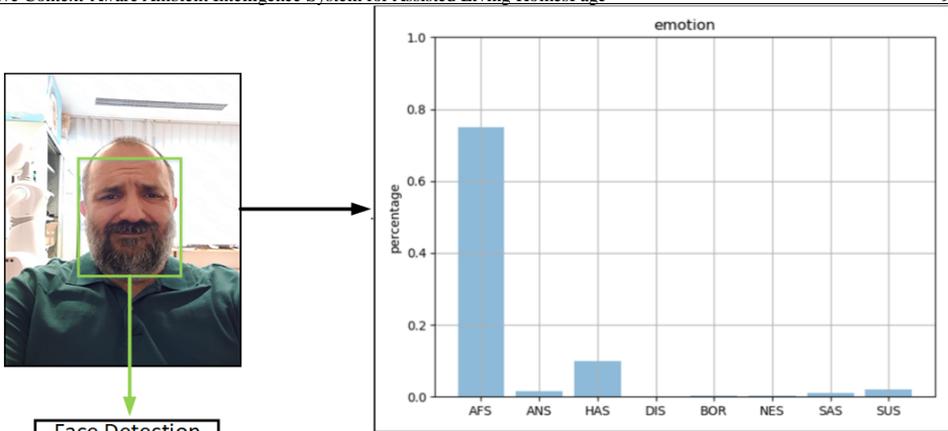

**Fig. 4.** Terrified emotion recognition.

## 6.2 Semantic Knowledge Augmentation and Types Learning

We utilized the first three stages of the context-aware knowledge acquisition 5-stages method using the ontologies approachpresented in [13] and designed initially to formulate new goals. Adaptation was required as the original system receives as input a new object and its type upfront, which is not available in our case. Therefore, in this adaptation, the first stage is applied, and then the second and third stages are applied for each type in the predefined set of object types. The ontological details are abstracted out in this paper, and a detailed description can be found in [13]. On the other hand, the output for our healthcare application domain is shown in Fig. 5.

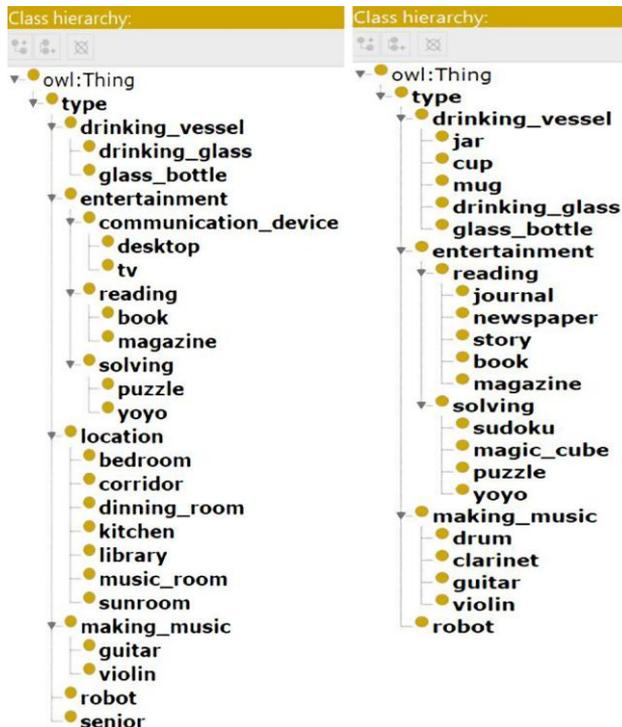

**Fig. 5.** OWL Representation of hierarchy of the predefined types and the augmented types.

Note the new types in Fig. 5: the types `jar`, `cup`, and `mug` under the superclass `drinking_vessel`; the types `journal`, `newspaper`, and `story` under the superclass `reading`; the types `sudoku` and `magic_cube` under the superclass `solving`; and the types `drum` and `clarinet` under the superclass `making_music`. The system will be trained on the augmented set of types to prepare it to recognize new objects that may comfort the senior in case of failure.

The system extracts the subclasses names of the augmented set of types and uses Google Images Download, a command-line Python program to download 200 images from Google of each subclass. We have developed a CNN composed of three convolutional layers for objects recognition [3, 32, 32]. The best results were obtained using three convolutional filters for each layer. The filter for each layer has the architecture [32, 32, 32, 32]. Finally, the fall rate was 0.2. Once the training is complete, the perception layer will be ready to observe new objects of the relevant types when the plan is executed. For `mug_id01` of type `mug for example`, the system can autonomously recognize the object, incorporate its related information, and integrate it within the planning task.

## 6.3 Planning, Acting, Failure Detection, and Monitoring

The planner provides a personalized plan that must reflect seniors' satisfaction on the AL home. The planner needs to consider the activities' durations and the IoT objects' availability to build the plan. Solving this problem requires using a planning system capable of dealing with durative actions to represent the duration of actions and hard goals according to seniors' emotions. For the implementation, among the few automated planners capable of handling temporal planning problems with time windows, we opted for the planner OPTIC because it handles the PDDL2.2.

The values of the planner's variables are read using the perception layer and compiled into a planning problem. The planner is called to generate plan $\pi_1$ (shown in Fig.6). $\pi_1$ consists of twelve actions for the robot `robot1` to move from `bedroom1` to `kitchen1`, carry and fill `glass_id05`, and then return to comfort terrified senior `senior1`.

```
a1  : 0.000   : (move robot1 bedroom1 corridor1)
a2  : 30.002  : (move robot1 corridor1 corridor2)
a3  : 80.004  : (move robot1 corridor2 dinning_room1)
a4  : 120.006 : (move robot1 dinning_room1 kitchen1)
a5  : 140.008 : (carry robot1 glass_id05 kitchen1)
a6  : 150.010 : (fill robot1 glass_id05 kitchen1)
a7  : 160.012 : (move robot1 kitchen1 dinning_room1)
a8  : 180.014 : (move robot1 dinning_room1 corridor2)
a9  : 220.016 : (move robot1 corridor2 corridor1)
a10 : 270.018 : (move robot1 corridor1 bedroom1)
a11 : 300.020 : (give robot1 senior1 glass_id05 bedroom1)
a12 : 310.022 : (soothed_received senior1 glass_id05)
```

**Fig. 6.** Temporal plan $\pi_1$.

To simulate actions execution monitoring while the robot is executing plan $\pi_1$, our simulator transforms $\pi_1$ and the planning task into an infrastructure called the timeline, similar to the work in [19]. In our case, however, the timeline is only a collection of chronologically ordered conditions that need to be satisfied in the observed world-state. The simulation starts at time $t = 0$ with the simulator's internal state equal to the initial state. Next, the idea is that the simulator advances through time in every execution step. Every execution step at time $t_i$ involves:

1) moving through time $t_i = t_i + stepsize$.
2) using the perception layer to read the observed state of the world.
3) checking if a failure is detected: a condition in the timeline is not satisfied in the observed state.
   a) The robot checks if it can recognize new objects of the augmented set of types and integrates them into the planning task along with their corresponding information.
   b) Goal generation is re-invoked to make sure the goal is still active.

c) The planning layer is re-invoked; the planning problem is reformulated with the new initial state set to the observed state and the newly integrated objects information and the goal. The planner is called to generate a new plan to achieve the goal. The acting layer is re-invoked.
4) If no failure is detected, the simulation continues from step 1 until all the conditions are monitored.

# 7. Validation

This section aims to demonstrate the validity of the following: (1) the behavior of our system using three simulated scenarios for fixing failures; (2) the proposed method for stress detection; (3) the proposed method for emotions recognition; and (4) the proposed method for classifying new objects.

## 7.1 Validity of Behavior of Our System

For simplicity, we show the behavior for one senior. In the first and second scenarios, the failure is repaired using the observed new objects, whereas it is fixed by suggesting an activity in the third scenario.

**Scenario 1.** Consider the healthcare scenario described in the concept design and the planning task sections. When perception detects a stressed senior, the robot recognizes his/her emotion as explained in the stress detection and emotions recognition section, which is terrified in this case. The system dynamically generates a new goal to comfort the terrified senior `(soothed senior1)` and adds the new goal to the planning problem.

Perception reads the origin places from the IoT environment objects (shown in Fig.1) and adds them to the planning problem: the robot is at `bedroom1` `(be robot1 bedroom1)`. The origin locations of each senior, e.g., `(be senior1 bedroom1)`. The origin locations for glasses at the kitchen, e.g., `(be glass_id05 kitchen1)`. The origin locations of instruments in the music room, `(be violin_id01 music_room1)` and `(be guitar_id01 music_room1)`. The origin locations of the entertainment objects in the library, `(be book_id01 library1)` and `(be magazine_id01 library1)`. The origin locations of communication devices (PCs and TV) in the library, `(be pc_id01 library1)`, `(be pc_id02 library1)`, and `(be tv_id01 library1)`. The availability of objects, `(available glass_id05)`, `(available glass_id06)`, `(available glass_id07)`, and `(available glass_id08)`. In addition to actions durations, such as the duration of movements between the locations, there are carrying and giving an item, filling drinking vessels, and suggesting an activity.

The planner is called to generate plan $\pi_1$ (Fig.6) consisting of 12 actions to comfort the terrified senior by providing the glass `glass_id05` filled with water. The system transforms $\pi_1$ to a timeline of events (chronologically ordered conditions). $\pi_1$ execution starts, the robot executes the first four actions correctly ($a_1$, $a_2$, $a_3$ and $a_4$), an exogenous event occurs, and all drinking vessels become unavailable. When the observed state is read at time 140.008, the system detects a violated condition `(available glass_id05)` required by $a_5$. Subsequently, a failure is detected.

The robot executes perception to read the new IoT objects values and recognize new objects. A new object of type `mug` is recognized. The object is named `mug_id01` and integrated into the planning problem along with its location `(be mug_id01 kitchen1)` and availability `(available mug_id01)`. The goal generation layer indicates that the senior is still terrified (the goal is still active). The planning problem is reformulated with the new observed state as the new initial state and the information of the new object `mug_id01` and the goal `(soothed senior1)`. The planner is called to fix the failure, and a new plan $\pi_1'$ is generated (Fig.7). The new object `mug_id01` is carried and filled, and then the robot moves to provide the terrified senior with the filled mug.

```
a₁ : 140.008 : (carry robot1 mug_id01 kitchen1)
a₂ : 150.010 : (fill robot1 mug_id01 kitchen1)
a₃ : 160.012 : (move robot1 kitchen1 dinning_room1)
a₄ : 180.014 : (move robot1 dinning_room1 corridor2)
a₅ : 220.016 : (move robot1 corridor2 corridor1)
a₆ : 270.018 : (move robot1 corridor1 bedroom1)
a₇ : 300.020 : (give robot1 senior1 mug_id01 bedroom1)
a₈ : 310.022 : (soothed_received senior1 mug_id01)
```

**Fig. 7.** Temporal plan $\pi_1'$.

**Scenario 2.** In this scenario, plan $\pi_2$ (shown in Fig.8) was generated to cheer up a sad senior, senior1 who is located at bedroom1. The robot moves from the senior's location to music_room, carries guitar_id01, and then moves back to cheer up the sad senior by giving guitar_id01.

```
a₁ : 0.000   : (move robot1 bedroom1 corridor1)
a₂ : 30.002  : (move robot1 corridor1 corridor2)
a₃ : 80.004  : (move robot1 corridor2 music_room1)
a₄ : 120.006 : (carry robot1 guitar_id01 music_room1)
a₅ : 130.008 : (move robot1 music_room1 corridor2)
a₆ : 170.010 : (move robot1 corridor2 corridor1)
a₇ : 220.012 : (move robot1 corridor1 bedroom1)
a₈ : 250.014 : (give robot1 senior1 guitar_id01 bedroom1)
a₉ : 260.016 : (cheer_received senior1 guitar_id01)
```

**Fig. 8.** Temporal plan $\pi_2$.

The robot executes the first three actions correctly ($a_1$, $a_2$, and $a_3$), an exogenous event occurs, and all music instruments become unavailable. When the observed state is read at 120.006, the system detects a violated condition (available guitar_id01) required by action $a_4$ in $\pi_2$. Subsequently, a failure is detected due to the discrepancy. Perception recognizes a new object of type clarinet. The object is named clarinet_id01 and is integrated into the planning task with its location (be clarinet_id01 music_room1) and availability (available clarinet_id01). The goal generation layer indicates that the senior is still sad. The planning problem is reformulated, and the planner is called to generate a new plan $\pi_2'$ (shown in Fig.9) that cheers up the sad senior by providing the clarinet_id01.

```
a₁ : 120.006 : (carry robot1 clarinet_id01 music_room1)
a₂ : 130.008 : (move robot1 music_room1 corridor2)
a₃ : 170.010 : (move robot1 corridor2 corridor1)
a₄ : 220.012 : (move robot1 corridor1 bedroom1)
a₅ : 250.014 : (give robot1 senior1 clarinet_id01 bedroom1)
a₆ : 260.016 : (cheer_received senior1 clarinet_id01)
```

**Fig. 9.** Temporal plan $\pi_2'$.

**Scenario 3.** In this scenario, plan $\pi_3$ (shown in Fig. 10) was generated to comfort a bored senior by providing the object book_id01 of type book.

During the execution, entertainment objects become unavailable. A discrepancy is detected due to book_id01. The system detects a failure. This time, the difference is that perception does not find any alternative objects in the execution environment; alternative objects cannot repair the failure. The planning task is reformulated. The planner is called from the state wherein the failure occurred to generate a new plan. The newly generated plan $\pi_3'$ (shown in Fig. 11) includes a suggested action to the bored senior to watch TV (a different course of actions).

```
a₁ : 0.000 : (move robot1 bedroom1 corridor1)
a₂ : 30.002 : (move robot1 corridor1 corridor2)
a₃ : 80.004 : (move robot1 corridor2 library1)
a₄ : 120.006 : (carry robot1 book_id01 library1)
a₅ : 130.008 : (move robot1 library1 corridor2)
a₆ : 170.010 : (move robot1 corridor2 corridor1)
a₇ : 220.012 : (move robot1 corridor1 bedroom1)
a₈ : 250.014 : (give robot1 senior1 book_id01 bedroom1)
a₉ : 260.016 : (entertainment_received senior1 book_id01)
```

**Fig. 10.** Temporal plan $\pi_3$.

```
a₁ : 0.000 : (suggest_entertainment2 robot1 senior1 tv_id01 bedroom1)
a₂ : 5.002 : (entertainment2_suggested senior1 tv_id01)
```

**Fig. 11.** Temporal plan $\pi_3'$

## 7.2 Validation of Stress Detection

Detecting stress is complicated mainly due to two factors; the determination as to which external elements are necessary to experience stress and the noise introduced when the senior moves the hand. The dataset we used was divided into two parts:80% for training and 20% for testing. The confusion matrix obtained by using the database as validation is presented in Table 1.

When validating using the data acquired by the wristband (shown in Table 2), a reduction in classification was observed, reducing it to 70%, due mainly to the two factors explained earlier. However, this is not an issue for our systemas we do not depend solely on stressbut instead continue to recognize the emotion and only then generate the goal.

The statistical data obtained from the stress classification process are analyzed to determine if our system performs the classificationcorrectly. The dataset's classification is 85% with average precision-recall of 0.8 but is 70% without the dataset with an average precision-recall of 0.65.

**Table 1.** Confusion matrix for the validation of stress detection using the dataset

|      |             | Predicted |             |
|------|-------------|-----------|-------------|
|      |             | Stressed  | Un-stressed |
| Real | Stressed    | 95.0      | 5.0         |
|      | Un-stressed | 11.0      | 89.0        |

**Table 2.** Confusion matrix for the validation of stress detection using the wristband

|      |             | Predicted |             |
|------|-------------|-----------|-------------|
|      |             | Stressed  | Un-stressed |
| Real | Stressed    | 75.0      | 25.0        |
|      | Un-stressed | 20.0      | 80.0        |

## 7.3 Validation of Emotions Recognition

A series of experiments were performed to validate the classification of emotions to determine classification accuracy. The robot captured some images wherein a person could be observed, trying to imitate a particular emotional state. These images were sent to the web service for analysis.

The result of these experiments is shownin Table 3, whereinthe confusion matrix obtained from our validation process can be seen. The accuracy obtained by the validation with our model is 93.6%. It is important to note that, in some emotions, some values are lower than other emotions due to the facial similarity in these emotions, such as sadness and anger.

**Table 3.** Confusion matrix for the validation of emotion classification

|      |           | Predicted |       |          |       |         |      |       |
|------|-----------|-----------|-------|----------|-------|---------|------|-------|
|      |           | **Terrified** | **Angry** | **Disgusted** | **Happy** | **Neutral** | **Sad** | **Bored** |
| Real | Terrified | **98.0**  | 3.1   | 0.0      | 2.0   | 1.3     | 0.0  | 0.87  |
|      | Angry     | 0.68      | **89.0** | 1.1   | 1.0   | 1.3     | 0.0  | 0.0   |
|      | Disgusted | 0.0       | 1.1   | **94.0** | 0.0   | 1.3     | 5.1  | 0.87  |
|      | Happy     | 0.34      | 2.1   | 3.3      | **94.0** | 2.6  | 5.1  | 1.7   |
|      | Neutral   | 0.68      | 1.1   | 1.1      | 3.0   | **91.0** | 1.7 | 0.0   |
|      | Sad       | 0.34      | 1.1   | 0.0      | 0.0   | 0.65    | **81.0** | 0.87 |
|      | Bored     | 0.34      | 2.1   | 0.0      | 0.0   | 2.0     | 6.8  | **96.0** |

## 7.4 Validation of Objects Classification

Not all the objects used in our system are found in typical datasets. Thus, we decided to build our 17-class dataset. For each class, a total of 200 images were downloaded. A MobileNet [14] network was used, and the classification results can be seen in the confusion matrix presented in Table 4. Some values of the objects in the confusion matrix were lower than other objects. The objects associated with these values, such as glass bottle, jar, and mug, have a certain resemblance. This is not an issue for our system as these objects belong to the same category—a drinking vessel—and they are intended to be used interchangeably. The lowest rating percentage (79.2%) is associated with the yo-yo, which is difficult to rate due to its shape.

## 8. Discussion

Abstractly, the system can be looked at as providing members of a target group with objects depending on their dynamically recognized emotions and following up on their status evolution in a dynamic execution environment. Since our deliberative function is domain-independent, the system can be further specialized depending on the target group. The system was not validated with elders in a real AL home, mainly due to the robotic platform we used and the safety regulations. We used the RobElf robotic platform, which does not have a clamp to pick up objects; to simulate actions that require a gripper, such as carrying a glass, it executes voice interaction. The system was presented to a group of users. They were all researchers in the Department of Computer Systems and Computation. For future research, we expect to validate and adapt the system following a user-centric methodology. We intend to survey impressions from real caregivers in a seniors' living home. We plan to acquire approval to conduct live laboratory demonstrations with senior citizens attended by health professionals. The demonstrations data must be analyzed to adjust system performance and avoid unforeseen inconvenient situations.

We opted for reactivity to reduce the information load in the current implementation. The system looks for alternative objects only when a failure occurs. On the other hand, the system can be proactive, looking for alternative objects every time the observed state is read. The newly integrated objects can be considered information overload when there are no failures, and they end up being unused. The reactive-proactive design detail is left to the system administrators to meet their environmental needs.

Mohannad Babli*, Jaime A Rincon, Eva Onaindia, Carlos Carrascosa, and Vicente Julia Page 15 / 18

**Table 4.** Confusion matrix for the validation of objects classification

| | | Predicted | | | | | | | | | | | | | | | | | |
|---|---|---|---|---|---|---|---|---|---|---|---|---|---|---|---|---|---|---|---|
| | | book | clarinet | cup | drinking_glass | drum | glass_bottle | guitar | jar | journal | magazine | magic_cube | mug | newspaper | puzzle | story | sudoku | violin | yo-yo |
| Real | book | **90** | 2.0 | 1.0 | 1.0 | 1.5 | 0.5 | 0.36 | 0.36 | 0.65 | 0.99 | 0.36 | 0.56 | 0.36 | 0.36 | 0.36 | 0.36 | 0.96 | 3.5 |
| | clarinet | 1.0 | **87** | 2.0 | 0.95 | 0.65 | 0.36 | 0.45 | 0.23 | 0.45 | 0.56 | 0.45 | 0 | 0.25 | 0.3 | 0 | 0.25 | 0.36 | 3.0 |
| | cup | 0.25 | 1.0 | **90** | 0.65 | 0.45 | 1.5 | 0.25 | 0.58 | 0.32 | 0.48 | 0.98 | 0.15 | 0.36 | 0.54 | 0.36 | 0.12 | 0.48 | 1.5 |
| | drinking_glass | 0.5 | 0.25 | 0.56 | **88** | 0.36 | 0.35 | 0.45 | 0.36 | 0.12 | 0.57 | 0.14 | 0.19 | 0.87 | 0.36 | 0.36 | 0.36 | 0.75 | 0.35 |
| | drum | 0.35 | 0.89 | 0.68 | 3.0 | **92** | 2.3 | 0.6 | 0.45 | 0.14 | 0.89 | 0.56 | 0.25 | 0.45 | 0.9 | 0.45 | 0.95 | 0.35 | 0.99 |
| | glass_bottle | 0.27 | 0.25 | 0.47 | 0.24 | 0.45 | **88** | 0.87 | 0.98 | 0.36 | 0.78 | 0.78 | 0.36 | 0.98 | 2.0 | 0.71 | 0.4 | 0.48 | 0.78 |
| | guitar | 0.5 | 0.36 | 0 | 0.45 | 0.25 | 0.69 | **90** | 0.78 | 0.98 | 0.65 | 0.9 | 0.98 | 0.78 | 0.36 | 0.56 | 0.84 | 3.0 | 2.5 |
| | jar | 0.5 | 0.48 | 0 | 0 | 1.5 | 0.95 | 2.3 | **88** | 1.5 | 0.56 | 0.65 | 0.31 | 0.36 | 0.97 | 0 | 3.0 | 0.34 | 3.6 |
| | journal | 1.5 | 0.9 | 0 | 0.92 | 0.9 | 0.35 | 0.23 | 2.3 | **90** | 0.36 | 0.78 | 2.5 | 0.29 | 0.45 | 0 | 0.45 | 0.14 | 0.58 |
| | magazine | 1.5 | 1.0 | 0 | 0.9 | 0.45 | 0.45 | 0.6 | 0.12 | 0.31 | **90** | 0.7 | 0.36 | 0 | 0.65 | 0 | 0.68 | 0.25 | 0.56 |
| | magic_cube | 0.68 | 0.96 | 0.56 | 0 | 0 | 0.67 | 0.78 | 0.36 | 0.98 | 0.45 | **88** | 0.45 | 0 | 0.31 | 0.3 | 0.99 | 0.89 | 0.14 |
| | mug | 0.25 | 0.78 | 0.9 | 0 | 0 | 0.68 | 0.89 | 0.69 | 0.36 | 0.87 | 0.98 | **88** | 0 | 0.45 | 0 | 0 | 0.35 | 0.23 |
| | newspaper | 0.55 | 0.95 | 0.65 | 0.24 | 0 | 0.69 | 0.45 | 0.97 | 0.93 | 0.98 | 1.5 | 2.5 | **95** | 1.8 | 0 | 0 | 0.35 | 0.69 |
| | puzzle | 0.1 | 1.0 | 0.35 | 0.35 | 0 | 0.99 | 0.63 | 0.45 | 0.45 | 1.0 | 0.36 | 0.36 | 0 | **88** | 0 | 0 | 0.99 | 0.99 |
| | story | 0.1 | 0.44 | 0.9 | 0.32 | 0 | 0.78 | 0.45 | 0.63 | 0.99 | 0 | 0.97 | 0.78 | 0 | 0.23 | **96** | 0 | 0.87 | 0.58 |
| | sudoku | 0.36 | 0.58 | 0.8 | 0.9 | 0 | 0.45 | 0.23 | 0.48 | 0.45 | 0 | 0.35 | 0.98 | 0 | 0.45 | 0 | **92** | 0.45 | 0.36 |
| | violin | 0.36 | 1.0 | 0.93 | 0.98 | 0 | 0.36 | 0.36 | 1.5 | 0.36 | 0 | 0.9 | 0.45 | 0 | 2.3 | 0 | 0 | **88** | 0.45 |
| | yo-yo | 0.98 | 0.56 | 0 | 1.5 | 0.99 | 0.43 | 0.1 | 0.69 | 0.25 | 0.36 | 0.36 | 0.32 | 0 | 0.07 | 1.3 | 0 | 1.5 | **79** |

In the validation section, we demonstrated the system's behavior with one senior for simplicity, but our implementation has multi-user capabilities. We utilized temporal planning limited time windows (timed initial literals) supported by PDDL2.2 to deal with multiple seniors suffering from negative emotions. When using such literals in actions conditions, time windows allow priorities. Comforting the terrified emotion is assigned a shorter time window than the ones assigned for comforting the sadness or the boredom emotions. The planner handles these constraints when generating the plan. In our current implementation, to serve the senior suffering from a negative emotion, durative actions are generated to provide a physical object. The work can be extended by adding instantaneous responses, e.g., turning on the TV and modifying lights intensity, as in [20], and further increasing multi-user capability as the robot is no longer a bottleneck resource.

Our system provides a contextualized explanation when it deviates from the expected behavior. The explanation is threefold: (1) based on the dynamically generated goal (reason for the original plan); (2) based on what went wrong (violated condition leading to the failure); and (3) based on the repair (reason for repairing using an alternative object or via a recommendation). In the current implementation, we have the assumption that a senior is comforted after a duration of time from receiving the object to soothe the negative emotion. If not comforted, the senior will be detected as stressed, and the process will start again. For future work, this is an area for improvement using explainable AIas people react differently.The system must learn why the suggested solution did not comfort the senior and adapt for future runs.

The solution to certain classes' similarity in the classification of emotions and objects is to add more images wherein a marked differentiation between the classes can be observed. Adding more images in which the actors emphasize anger and sad emotions would allow the system to identify these emotions correctly. Similarly, introducing images from different perspectives or rotating the images can improve classification for the glass bottle, the jar, and the cup. The noise to the wristband when the senior moves the hand can lead to false positives. A possible solution is to incorporate a breathing frequency measurement in the wristband.Thus, the system would have one more variable to infer if the user is exhibiting a state of stress. Another possible solution is to calculate the average of the HR measurements over a period of time and send the result to the system for classification.

## 9. Conclusion

The use of AmI and AL systems is increasing in line with the growing agingpopulation. Stress is a significant cause of human illnesses directly affecting the emotional state. Literature was reviewed to identify the convergence between deliberation and AmI latest healthcare trends and the lack of a balanced deliberative function. As the contribution of this paper, it describes an integral deliberation function architecture that provides comfortto stressed seniors suffering from negative emotions in an AL environment. The integral deliberation function makes a balanced synthesis of data and knowledge, learning, perception, dynamic goal generation, planning, acting, and monitoring functionalities. Such functionalities are required to detect, address, and follow up on the senior's status evolution in a dynamic execution environment. The five layers constituting our intelligent deliberation pillars are learning and knowledge augmentation, perception, goal generation, planning, acting, and monitoring. Our contribution's differential value lies in the integral view of the architecture: (1) integrates learning, perception, and dynamic goal generation to achieve context-aware human-robot interaction;(2) integrates cognitive knowledge augmentation, learning, and perception to achieve context-awareness towardthe execution environment;and (3)integrates perception, planning, acting, and monitoring to achieve planning capabilities and reactivity when handling failures.

The behavior and the approach's validity were demonstrated in three experimental case studies in a simulatedAL home scenario. Moreover, the proposed methods for stress detection, emotions, and objects recognition were validated to show the classification accuracy. The validation demonstrated that the proposed deliberation function has effectively achieved its deliberative objectives in the ALhome:perceptiveness and context-aware dynamic human-robot interaction; planning, acting, and monitoring capabilities to achieve reactiveness; and environment context-awareness to adapt to the change in the environment. For future research, we are working on alternative plan repair mechanisms to provide diverse plans instead of delegating to the planner. The convergence between deliberation and ambient intelligence could lead to new interdisciplinary research models and new assistive systems for seniors' healthcare.

## Acknowledgements

Not applicable.

## Author's Contributions

Conceptualization, MB, EO. Investigation and methodology, MB, EO, JAR.Supervision, EO, CC, VJ. Writing of the original draft, MB, JAR.Writing of the review and editing, MB.All authors read and approved the final manuscript.

## Funding

This work is supported by the Spanish MINECO Project (No. TIN2017-88476-C2-1-R) and the Universitat Politecnica de Valencia Research (Grant No. PAID-10-19).

## Competing Interests

The authors declare that they have no competing interests.